\documentclass[journal]{IEEEtran}
\usepackage{multirow}
\usepackage{bm}
\usepackage{bbm}
\usepackage{algorithm}
\usepackage{algorithmic}
\usepackage{subfig} 
\usepackage{amsmath,amssymb}
\usepackage[switch]{lineno}
\usepackage{graphicx}
\usepackage[numbers,sort]{natbib} 
\usepackage{comment}
\usepackage{booktabs}

\DeclareMathOperator*{\argmax}{arg\,max}

\newcommand{\std}[1]{\tiny$\pm${#1}}
\newcommand{\codecmt}[1]{\textit{// #1}}

\newcommand{\tableCellHeight}{1}
\newcommand{\tabstyle}[1]{
  \setlength{\tabcolsep}{#1}
  \renewcommand{\arraystretch}{\tableCellHeight}
  \centering
  \small
}

\ifCLASSINFOpdf
\else
\fi
\usepackage{url}

\begin{document}
%
\title{Domain Adaptive Ensemble Learning}
%
%
%

\author{Kaiyang~Zhou,
        Yongxin~Yang,
        Yu Qiao,
        and~Tao~Xiang.
\thanks{K.~Zhou is with Nanyang Technological University, Singapore.}
\thanks{Y.~Yang and T.~Xiang are with the University of Surrey, UK.}
\thanks{Y.~Qiao is with the Shenzhen Institutes of Advanced Technology, Chinese Academy of Sciences, China.}
}

%
%

\markboth{Journal of \LaTeX\ Class Files,~Vol.~14, No.~8, August~2015}%
{Shell \MakeLowercase{\textit{et al.}}: Bare Demo of IEEEtran.cls for IEEE Journals}
%



\maketitle

\begin{abstract}
The problem of generalizing deep neural networks from multiple source domains to a target one is studied under two settings: When unlabeled target data is available, it is a multi-source unsupervised domain adaptation (UDA) problem, otherwise a domain generalization (DG) problem. We propose a unified framework termed domain adaptive ensemble learning (DAEL) to address both problems. A DAEL model is composed of a CNN feature extractor shared across domains and multiple classifier heads each trained to specialize in a particular source domain. Each such classifier is an expert to its own domain but a non-expert to others. DAEL aims to learn these experts collaboratively so that when forming an ensemble, they can leverage complementary information from each other to be more effective for an unseen target domain. To this end, each source domain is used in turn as a pseudo-target-domain with its own expert providing supervisory signal to the ensemble of non-experts learned from the other sources. To deal with unlabeled target data under the UDA setting where real expert does not exist, DAEL uses pseudo labels to supervise the ensemble learning. Extensive experiments on three multi-source UDA datasets and two DG datasets show that DAEL improves the state of the art on both problems, often by significant margins.
\end{abstract}


%
\IEEEpeerreviewmaketitle

\section{Introduction}
%
%
%
%
\IEEEPARstart{D}{eep} neural networks trained with sufficient labeled data typically perform well when the test data follows a similar distribution as the training data. However, when the test data distribution is different, neural networks often suffer from significant performance degradation. Such a problem is common to machine learning models and is often referred to as domain shift~\cite{zhou2021domain} (or distribution shift). To overcome the domain shift problem, two related areas have been studied extensively, namely \emph{unsupervised domain adaptation} (UDA)~\cite{ganin2015unsupervised,long2015learning,nguyen2015dash,hou2016unsupervised,cvpr18MCD,lu2018embarrassingly,li2019locality,Xu_2019_ICCV} and \emph{domain generalization} (DG)~\cite{zhou2020learning,muandet2013domain,li2017deeper,shankar2018generalizing,zhou2020deep,zhou2021mixstyle}. UDA aims to adapt a model from a labeled source domain to an unlabeled target domain. In contrast, DG aims to learn a model only from source data typically gathered from multiple distinct but related domains, and the model is directly deployed in a target domain without any fine-tuning or adaptation steps.

Early UDA work focuses on single-source scenarios. Recently, \emph{multi-source} UDA~\cite{hoffman2018algorithms,xu2018deep,iccv19DomainNet} has started to attract more attention, thanks to the introduction of large-scale multi-domain datasets such as DomainNet~\cite{iccv19DomainNet}. In contrast, having multiple source domains has been the default setting for most DG methods from much early on~\cite{muandet2013domain}. This is understandable: without the guidance from target domain data, DG models rely on the diversity of source domain to learn generalizable knowledge. This paper focuses on the multi-source setting for both problems.

\begin{figure*}[t]
    \centering
    \includegraphics[width=.8\textwidth]{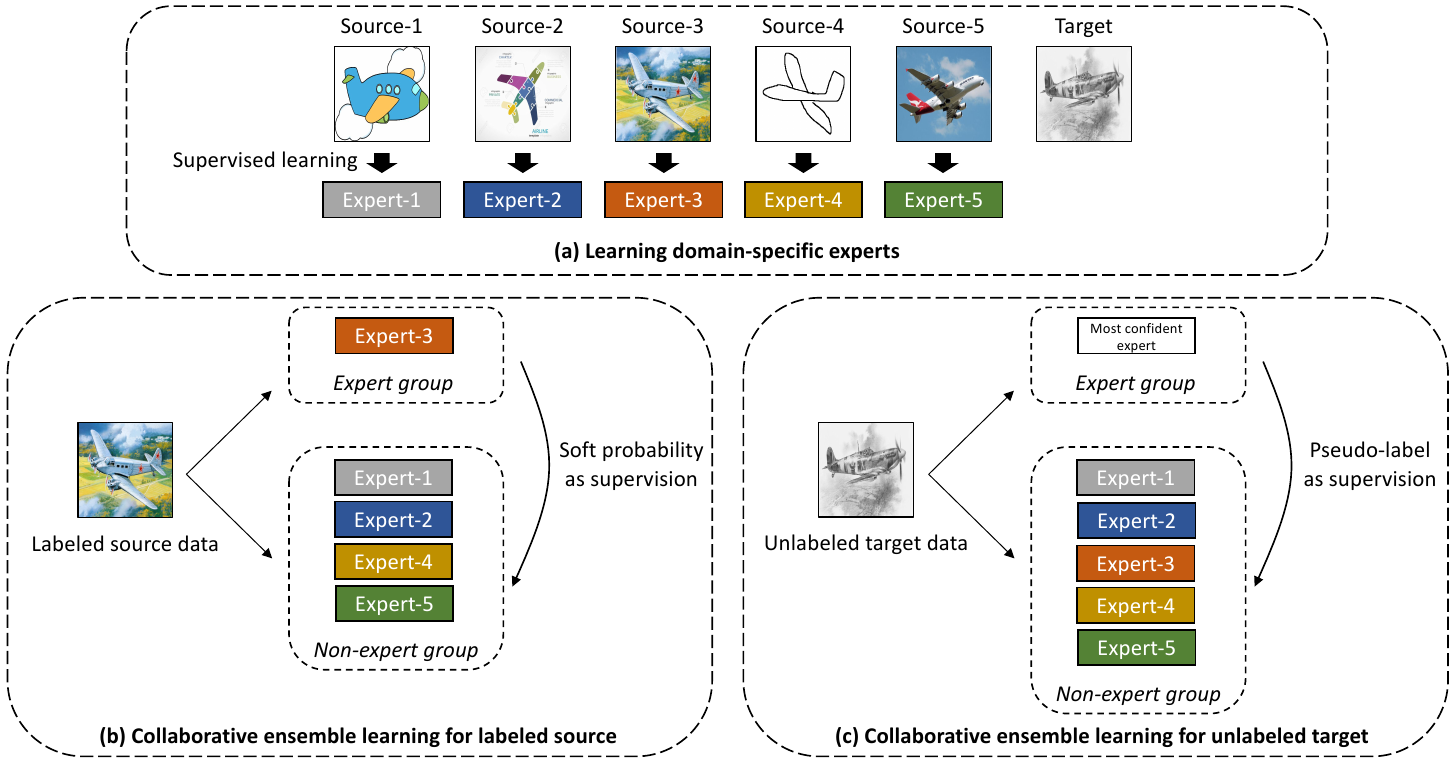}
    \caption{Overview of domain adaptive ensemble learning (DAEL).}
    \label{fig:concept}
\end{figure*}

How can multiple source domains be exploited to help generalization? Many DG methods~\cite{motiian2017unified,ghifary2017scatter,li2018mmdaae} aim to learn a domain-invariant feature representation or classifier across the source domains, in the hope that it would also be invariant to domain shift brought by the target domain. However, there is an intrinsic flaw in this approach, that is, when the source domains become more diverse, learning a domain-invariant model becomes more difficult. This is because each domain now contains much domain-specific information. Simply removing the information may be detrimental to model generalization because such information could potentially be useful for a target domain, especially when combined across different source domains. An example can be found in Fig.~\ref{fig:concept}(a) where the only thing in common of the five source domains for the airplane class seems to be shape. However, texture information is also useful for object recognition in the target sketch domain, which we want to maintain in the learned classifier. Existing multi-source UDA methods, on the other hand, attempt to align the data distribution of the target domain with each source domain individually~\cite{zhao2018adversarial,xu2018deep,iccv19DomainNet} or by means of a hard~\cite{li2018extracting} or soft~\cite{hoffman2018algorithms} domain selector. Again, Fig.~\ref{fig:concept}(a) suggests that aligning the target domain to each individual source domain is not only difficult but could also be counterproductive due to drastic variations among source domains.

In this paper, we propose a novel unified framework for both multi-source DG and UDA based on the idea of collaborative ensemble learning. Our framework, termed \emph{domain adaptive ensemble learning} (DAEL), takes a very different approach from previous work. Specifically, each domain is used to learn a model that is specialized in that domain (see Fig.~\ref{fig:concept}(a)). We call it a domain expert---a relative term as an expert to a specific source domain would be a non-expert to all other source domains as well as the target domain. The key idea of DAEL is to learn these experts collaboratively so that when forming an ensemble, they can leverage complementary information to better tackle the target domain.

To realize the DAEL framework for a UDA or DG model, a number of issues need to be addressed. (1) Scalability: Training an ensemble of models instead of a single model means higher computational cost. To solve this problem, we design a DAEL model as a deep multi-expert network consisting of a shared convolutional neural network (CNN) feature extractor and multiple classifier heads. Each head is trained to classify images from a particular source domain. Therefore, different heads learn different patterns from the shared features for classification. (2) Training: Since the target domain data is either non-existent (for DG) or has no label (for UDA), there is no target domain expert to provide supervisory signal for the source domain expert ensemble. To overcome this, each source domain is used in turn as a pseudo-target-domain with its own expert providing supervisory signal to the ensemble of non-experts learned from the other sources (see Fig.~\ref{fig:concept}(b)). For unlabeled target data under the UDA setting where real expert does not exist, DAEL uses as pseudo-label the most confident estimation among all experts and train the ensemble to fit the pseudo-label (see Fig.~\ref{fig:concept}(c)). (3) How to measure the effectiveness of a non-expert ensemble w.r.t.~an expert: Inspired by consistency regularization (CR)~\cite{sajjadi2016regularization,laine2016temporal} used in semi-supervised learning, the ensemble's effectiveness is measured by how close its prediction is to that of an expert when both are fed with a data point from the expert's domain. To amplify the regularization effect brought by CR, we use weak and strong augmentation for input to an expert and a non-expert ensemble respectively. Such a strategy has been shown useful in recent semi-supervised learning methods~\cite{xie2019unsupervised,berthelot2020remixmatch,sohn2020fixmatch}. Once these three issues are addressed, we have a simple but effective solution to both UDA and DG. By sending supervisory signal to an ensemble rather than each individual, different domain-specific experts are allowed to exploit complementary domain-specific information from each other, resulting in a more domain-generalizable ensemble.

We summarize our \textbf{contributions} as follows.
\textbf{(1)} We present a novel framework called domain adaptive ensemble learning (DAEL), which improves the generalization of a multi-expert network by explicitly training the ensemble to solve the target task.
\textbf{(2)} A realization of DAEL is formulated which provides a simple yet effective solution for both multi-source UDA and DG, unlike previous methods that only tackle one of them.
\textbf{(3)} We define miniDomainNet, a reduced version of DomainNet~\cite{iccv19DomainNet} to allow fast prototyping and experimentation. For benchmarking, a unified implementation and evaluation platform of all compared methods is created, called \texttt{Dassl.pytorch}, which has been made publicly available.\footnote{\url{https://github.com/KaiyangZhou/Dassl.pytorch}.}
\textbf{(4)} We demonstrate the effectiveness of DAEL on three multi-source UDA datasets and two DG datasets where DAEL outperforms the current state of the art by a large margin (see Table~\ref{tab:msUDA_results} \&~\ref{tab:DG_results}).

\section{Related Work} \label{sec:related_work}

\textbf{Unsupervised domain adaptation.}
Motivated by the seminal theory work by Ben-David et al.~\cite{ben2010theory}, numerous UDA methods seek to reduce distribution discrepancy between source and target features using some distance metrics, such as maximum mean discrepancy~\cite{long2015learning,long2016unsupervised,long2017deep}, optimal transport~\cite{bhushan2018deepjdot,balaji2019normalized}, and graph matching~\cite{das2018graph,yang2019cross,pilanci2020domain}. Inspired by generative adversarial network (GAN)~\cite{goodfellow2014generative}, several methods~\cite{ganin2015unsupervised,tzeng2017adversarial,zhang2018collaborative,long2018conditional,chadha2019improved} additionally train a domain discriminator for feature alignment. GAN has also been exploited for pixel-level domain adaptation where target images are synthesized via image translation/generation~\cite{bousmalis2017unsupervised,li2020generating}. Instead of aligning the coarse marginal distribution, recent alignment methods have shown that fine-grained alignment such as aligning class centroids~\cite{xie2018learning,kang2019contrastive,deng2019cluster,kang2020contrastive} or using task-specific classifiers~\cite{cvpr18MCD,lee2019sliced,iccv19MME} can give a better adaptation performance.

The multi-source UDA methods are more related to our work because of the same problem setting. Several works~\cite{zhao2018adversarial,xu2018deep,iccv19DomainNet} extend the domain alignment idea to multi-source UDA by considering all possible source-target pairs. Kang et al.~\cite{kang2020contrastive} propose contrastive adaptation network where a contrastive domain discrepancy loss is minimized for samples from the same class but of different domains while maximized for samples from different classes. Relationships between each source and the target are learned by Li et al.~\cite{li2018extracting} and only the target-related sources are kept for model learning. Hoffman et al.~\cite{hoffman2018algorithms} compute distribution-based weights for combining source classifiers. Our model architecture---a shared feature extractor and multiple domain-specific classifiers---is similar to M$^3$SDA~\cite{iccv19DomainNet}. However, DAEL is very different in that different domain-specific classifiers are learned \emph{collaboratively} where each source domain is used in turn as a \emph{pseudo-target-domain} to train the ensemble.

\textbf{Domain generalization.}
Many DG methods follow the idea of distribution alignment originated from the UDA community to learn domain-invariant features through minimizing in-between-source distances~\cite{motiian2017unified,ghifary2017scatter,li2018mmdaae}. Data augmentation is another popular research direction where the motivation is to avoid overfitting to source data. This can be achieved by, for example, adding adversarial gradients to the input~\cite{shankar2018generalizing,volpi2018generalizing}, learning data generation networks~\cite{zhou2020deep,zhou2020learning}, or mixing instance-level feature statistics~\cite{zhou2021mixstyle}. Meta-learning has also been investigated for learning domain-generalizable neural networks~\cite{li2018learning,balaji2018metareg,dou2019domain,li2019episodic}. Different from existing DG methods that mostly train a single classifier, our work for the first time introduces collaborative ensemble learning to mine domain-specific information using domain-specific classifiers. Although our pseudo-target-domain idea is related in spirit to meta-learning, no episodic training is required in DAEL, which makes the training procedure much simpler. We refer readers to Zhou et al.~\cite{zhou2021domain} for a comprehensive survey in DG.

\textbf{Ensemble methods}
have been extensively researched in the machine learning community~\cite{zhou2012ensemble}. The principle is to train multiple learners for the same problem and combine them for inference. Such technique has also been widely used in competitions like ILSVRC~\cite{deng2009imagenet} where multiple CNNs are trained and combined to improve the test performance~\cite{krizhevsky2012imagenet,he2016deep}. In this work, to prompt the emergence of generalizable features, we learn an ensemble of classifiers (experts) in a collaborative way---using each individual expert to supervise the learning of the non-expert ensemble.

\section{Methodology} \label{sec:method}

\begin{figure*}[t]
    \centering
    \includegraphics[width=.8\textwidth]{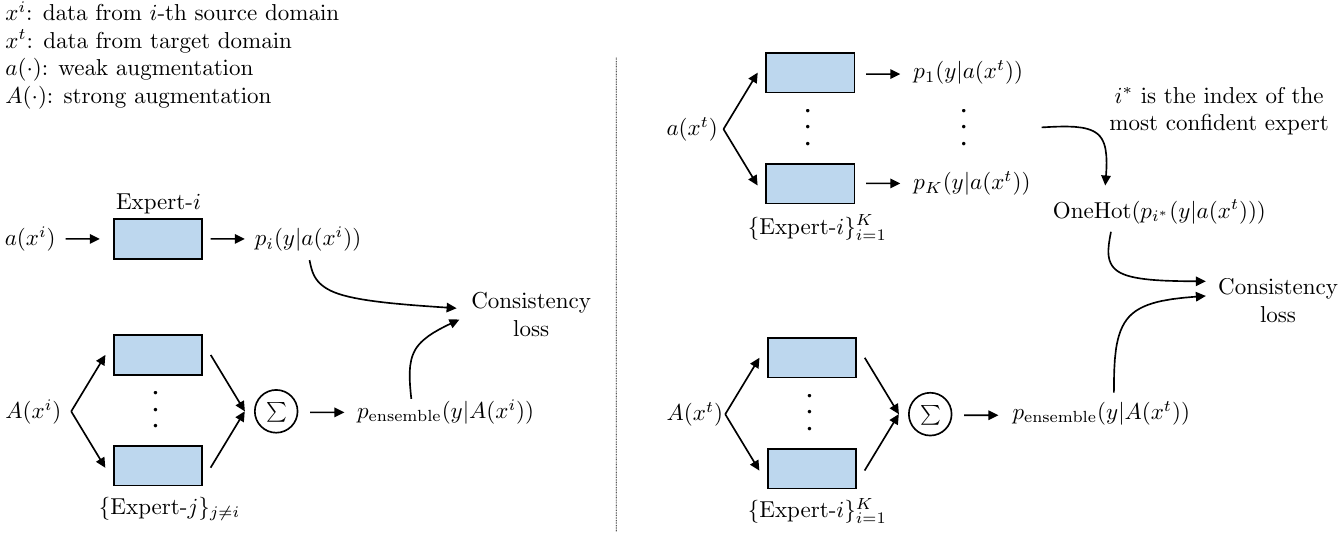}
    \caption{Illustration of domain adaptive ensemble learning. Left: collaborative learning using source domains. Right: collaborative learning using unlabeled target domain via pseudo-labeling. The ensemble of all experts is used for testing. Gradients are only back-propagated through the ensemble prediction path.
    }
    \label{fig:alg_overview}
\end{figure*}

\textbf{Problem definition.}
Given a labeled training dataset collected from $K$ source domains, $\mathcal{D}_S = \{ \mathcal{D}_1, ..., \mathcal{D}_K \}$, we aim to learn a model that can generalize well to a target domain $\mathcal{D}_T$. If the unlabeled target data is available during training, it is a multi-source unsupervised domain adaptation (UDA) problem~\cite{iccv19DomainNet}, otherwise a domain generalization (DG) problem~\cite{muandet2013domain}. Our method addresses these two problems in a unified framework.

\textbf{Model.}
We aim to learn a multi-expert model, denoted by $\{ E_i \}_{i=1}^K$, with each expert $E_i$ specializing in a particular source domain $\mathcal{D}_i$. For clarity, $E_i$ is called an expert to $\mathcal{D}_i$ but a non-expert to $\{ \mathcal{D}_j \}_{j \neq i}$. The ensemble prediction for an image $x$ is used at test time, i.e. $p(y|x) = \frac{1}{K} \sum_{i=1}^K E_i (x)$. In implementation, the multi-expert model shares a CNN backbone for feature extraction, followed by domain-specific classification heads. To allow the ensemble to better exploit complementary information between experts, we propose domain adaptive ensemble learning (DAEL). The main idea of DAEL is to strengthen the ensemble's generalizability by simulating how it is tested---using an expert's output to supervise the learning of ensemble of non-experts. This is realized by consistency regularization (CR) training, as shown in Fig.~\ref{fig:alg_overview}. To amplify the regularization effect, we follow Sohn et al.~\cite{sohn2020fixmatch} to use weak and strong augmentations, denoted by $a(\cdot)$ and $A(\cdot)$ respectively. Specifically, weak augmentation, which corresponds to simple flip-and-shift transformations, is used for pseudo-label generation; strong augmentation, which induces stronger noises like rotation and shearing, is used for ensemble prediction.

\textbf{Domain-specific expert learning.}
Next, we detail the DAEL training procedure, starting with how each expert is trained to be domain-specific. Let $H(\cdot, \cdot)$ denote cross-entropy between two probability distributions, the loss function for domain-specific expert learning is
\begin{equation} \label{eq:L_ce}
\mathcal{L}_{ce} = \frac{1}{K} \sum_{i=1}^K \mathbb{E}_{x^i, y(x^i) \sim \mathcal{D}_i} [ H(y(x^i), E_i(a(x^i))) ],
\end{equation}
where $y(x^i)$ is the one-hot label of $x^i$; the expectation is implemented by mini-batch sampling (same for the following equations).

\textbf{Collaborative ensemble learning using source domain data.}
Given an image $x^i$ from the $i$-th source domain (treated as a \emph{pseudo-target-domain}), the idea is to use as target the corresponding expert's prediction for the weakly augmented image, $E_i(a(x^i))$, and encourage the ensemble prediction of non-experts from other source domains for the strongly augmented image, $\frac{1}{K-1}\sum_{j \neq i} E_j(A(x^i))$, to be close to the target. Such a design explicitly teaches the ensemble how to handle data from unseen domains (mimicked by strong augmentation and guided by a pseudo-target-domain expert), thus improving the robustness to domain shift. Formally, the loss is defined as the mean-squared error (MSE) between the two outputs:\footnote{We chose MSE over KL divergence because the former led to a slightly higher performance.}
\begin{equation} \label{eq:L_cr}
\mathcal{L}_{cr} = \frac{1}{K} \sum_{i=1}^K \mathbb{E}_{x^i \sim \mathcal{D}_i} \big[ \| E_i(a(x^i)) - \frac{1}{K-1}\sum_{j \neq i} E_j(A(x^i)) \|^2 \big].
\end{equation}

\textbf{Collaborative ensemble learning using unlabeled target data.}
Given a weakly augmented target domain image $a(x^t)$, we first ask each source-expert to produce a class probability distribution, $p_i(y|a(x^t)) = E_i(a(x^t))$, and select as pseudo-label the most confident expert's prediction based on their maximum probability, $\argmax(p_{i^*})$, where $i^*$ is the index of the most confident expert. This is inspired by the observation that correct predictions are usually confident with peaked value on the predicted class~\cite{sohn2020fixmatch}. Then, we force the ensemble prediction of all source-experts for the strongly augmented image, $\bar{E}(A(x^t)) = \frac{1}{K}\sum_{i=1}^K E_i(A(x^t))$, to fit the one-hot pseudo-label $\hat{y}(x^t) = \argmax(p_{i^*})$.\footnote{For simplicity we assume $\argmax$ converts soft probability to one-hot encoding.} The loss is defined as
\begin{equation} \label{eq:L_u}
\mathcal{L}_{u} = \mathbb{E}_{x^t \sim \mathcal{D}_T} [\mathbbm{1}(\max(p_{i^*}) \geq \epsilon) H (\hat{y}(x^t), \bar{E}(A(x^t))) ],
\end{equation}
where $\epsilon$ is a confidence threshold (fixed to $0.95$ in this paper).

Eq.~\eqref{eq:L_u} can be viewed as a combination of CR and entropy minimization~\cite{grandvalet2005semi} because the conversion from soft probability to one-hot encoding essentially reduces the entropy of the class distribution. The confidence threshold provides a curriculum for filtering out less confident (unreliable) pseudo labels in the early training stages~\cite{iclr18SE,sohn2020fixmatch}.

\textbf{The full learning objective}
is a weighted sum of Eq.~\eqref{eq:L_ce}, \eqref{eq:L_cr} and \eqref{eq:L_u},
\begin{equation} \label{eq:L_full}
\mathcal{L} = \mathcal{L}_{ce} + \mathcal{L}_{cr} + \lambda_{u} \mathcal{L}_{u},
\end{equation}
where $\lambda_{u}$ is a hyper-parameter for balancing the weighting between $\mathcal{L}_{u}$ and the losses for the labeled source domains. For multi-source UDA, DAEL uses Eq.~\eqref{eq:L_full}. For DG, $\mathcal{L}_{u}$ is removed due to the absence of target domain data. DAEL not only provides a unified solution to the two problems, but is also very easy to implement (see Appendix~\ref{appx:pseudo_code} for pseudo-code).

\textbf{Gradient analysis.}
To understand the benefit of collaborative learning (i.e., $\|\frac{1}{K}\sum_i p_i - p^* \|^2$) against individual learning (i.e., $\frac{1}{K}\sum_i \|p_i - p^*\|^2$) where $p^*$ denotes the target, we analyze their gradients with respect to a single expert's output $p_i$. For collaborative learning, we have $\Delta p_i = \frac{2}{K} (\frac{1}{K}(p_i + \sum_{j \neq i} p_j) - p^*)$. For individual learning, we have $\Delta p_i = \frac{2}{K} ( p_i - p^*)$. It is clear that collaborative learning updates an expert by combining information from other experts, which facilitates the exploitation of complementary information.\footnote{The same conclusion can be drawn when using KL divergence as the objective.} Table~\ref{tab:col_vs_ind} further confirms the advantage of collaborative learning.

\textbf{Relation to knowledge distillation.}
DAEL is similar to knowledge distillation (KD)~\cite{hinton2015distilling} in the sense that the teacher-student training is used. However, in DAEL the boundary between teacher and student is blurred because each student can become a teacher when the input data come from its domain of expertise. Moreover, the collaborative ensemble learning strategy is specifically designed for dealing with multi-domain data---it encourages different experts to learn complementary information such that the ensemble is more generalizable to unseen domains.

\section{Experiments} \label{sec:experiments}

\subsection{Experiments on Domain Adaptation}

\begin{figure*}[t]
    \centering
    \includegraphics[width=.75\textwidth]{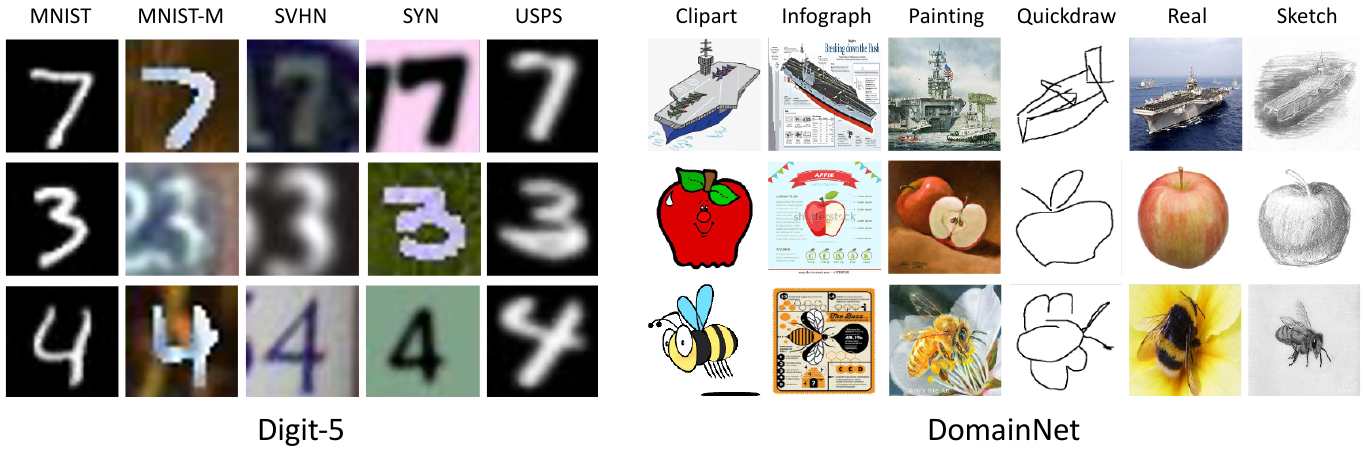}
    \caption{Example images from Digit-5 and DomainNet.}
    \label{fig:examples_digit5_domainnet}
\end{figure*}

\textbf{Datasets.}
(1) \textbf{Digit-5} consists of five different digit recognition datasets, which are MNIST~\cite{lecun1998mnist}, MNIST-M~\cite{ganin2015unsupervised}, USPS, SVHN~\cite{netzer2011svhn} and SYN~\cite{ganin2015unsupervised}. We follow the same setting as in M$^3$SDA~\cite{iccv19DomainNet} for experimentation. See Fig.~\ref{fig:examples_digit5_domainnet} left for example images.
(2) \textbf{DomainNet}~\cite{iccv19DomainNet} is a recently introduced benchmark for large-scale multi-source domain adaptation. It has six domains (Clipart, Infograph, Painting, Quickdraw, Real and Sketch) and 0.6M images of 345 classes.\footnote{We have noticed that the `t-shirt' class (index 327) is excluded from Painting's training set (see the official \texttt{painting\_train.txt} file), but is included in the test set, which could affect the performance.} See Fig.~\ref{fig:examples_digit5_domainnet} right for example images.
(3) The full DomainNet requires considerable computing resources for training,\footnote{It usually takes several GPU days for training a deep model on the full DomainNet.} preventing wide deployment and extensive ablative studies. Inspired by the miniImageNet dataset~\cite{vinyals2016matching} that has been widely used in the few-shot learning community, we propose \textbf{miniDomainNet}, which takes a subset of DomainNet and uses a smaller image size ($96 \times 96$). As noted by Saito et al.~\cite{iccv19MME} that the labels of some domains and classes are very noisy in the original DomainNet, we follow them to select four domains and 126 classes. As a result, miniDomainNet contains 18,703 images of Clipart, 31,202 images of Painting, 65,609 images of Real and 24,492 images of Sketch. In general, miniDomainNet maintains the complexity of the original DomainNet, reduces the requirements for computing resources and allows fast prototyping and experimentation.

\begin{table*}[t]
\caption{Comparing DAEL with state of the art on multi-source UDA datasets.}
\label{tab:msUDA_results}
\centering
\subfloat[Digit-5.]{
\tabstyle{5pt}
\begin{tabular}{l | c c c c c | c}
\toprule
Method & MNIST-M & MNIST & USPS & SVHN & SYN & Avg \\
\midrule
Oracle & 95.36\std{0.15} & 99.50\std{0.08} & 99.18\std{0.09} & 92.28\std{0.14} & 98.69\std{0.04} & 97.00 \\ \midrule
Source-only & 68.08\std{0.39} & 99.06\std{0.05} & 97.20\std{0.48} & 84.56\std{0.36} & 89.87\std{0.32} & 87.75 \\
DCTN~\cite{xu2018deep} & 76.20\std{0.51} & 99.38\std{0.06} & 94.39\std{0.58} & 86.37\std{0.54} & 86.78\std{0.31} & 88.63 \\
DANN~\cite{ganin2016domain} & 83.44\std{0.12} & 98.46\std{0.07} & 94.19\std{0.31} & 84.08\std{0.60} & 92.91\std{0.23} & 90.61 \\
CMSS~\cite{yang2020curriculum} & 75.30\std{0.57} & 99.00\std{0.08} & 97.70\std{0.13} & 88.40\std{0.54} & 93.70\std{0.21} & 90.80 \\
MCD~\cite{cvpr18MCD} & 80.65\std{0.51} & 99.22\std{0.08} & 98.32\std{0.07} & 81.87\std{0.72} & 95.42\std{0.04} & 91.09 \\
SE~\cite{iclr18SE} & 80.16\std{0.48} & 99.41\std{0.06} & 98.87\std{0.08} & 86.15\std{0.76} & 96.44\std{0.34} & 92.20 \\
MME~\cite{iccv19MME} & 83.07\std{0.57} & 99.35\std{0.03} & 98.64\std{0.17} & 86.40\std{0.41} & 95.78\std{0.15} & 92.65 \\
M$^3$SDA~\cite{iccv19DomainNet} & 82.15\std{0.49} & 99.38\std{0.07} & \textbf{98.71}\std{0.12} & 88.44\std{0.72} & 96.10\std{0.10} & 92.96 \\
DAEL (\emph{ours}) & \textbf{93.77}\std{0.12} & \textbf{99.45}\std{0.02} & 98.69\std{0.79} & \textbf{92.50}\std{0.15} & \textbf{97.91}\std{0.03} & \textbf{96.47} \\
\bottomrule
\end{tabular}
\label{tab:results_digit5}
}
\\
\subfloat[DomainNet.]{
\tabstyle{3pt}
\begin{tabular}{l | c c c c c c | c}
\toprule
Method & Clp & Inf & Pnt & Qdr & Rel & Skt & Avg \\
\midrule
Oracle~\cite{iccv19DomainNet} & 69.3\std{0.37} & 34.5\std{0.42} & 66.3\std{0.67} & 66.8\std{0.51} & 80.1\std{0.59} & 60.7\std{0.48} & 63.0 \\ \midrule
Source-only~\cite{iccv19DomainNet} & 47.6\std{0.52} & 13.0\std{0.41} & 38.1\std{0.45} & {13.3}\std{0.39} & 51.9\std{0.85} & 33.7\std{0.54} & 32.9 \\
DANN~\cite{ganin2016domain} & 45.5\std{0.59} & 13.1\std{0.72} & 37.0\std{0.69} & 13.2\std{0.77} & 48.9\std{0.65} & 31.8\std{0.62} & 32.6 \\
DCTN~\cite{xu2018deep} & 48.6\std{0.73} & 23.5\std{0.59} & 48.8\std{0.63} & 7.2\std{0.46} & 53.5\std{0.56} & 47.3\std{0.47} & 38.2 \\
MCD~\cite{cvpr18MCD} & 54.3\std{0.64} & 22.1\std{0.70} & 45.7\std{0.63} & 7.6\std{0.49} & 58.4\std{0.65} & 43.5\std{0.57} & 38.5 \\
M$^3$SDA~\cite{iccv19DomainNet} & 58.6\std{0.53} & 26.0\std{0.89} & 52.3\std{0.55} & 6.3\std{0.58} & 62.7\std{0.51} & 49.5\std{0.76} & 42.6 \\
CMSS~\cite{yang2020curriculum} & 64.2\std{0.18} & \textbf{28.0}\std{0.20} & 53.6\std{0.39} & \textbf{16.0}\std{0.12} & 63.4\std{0.21} & 53.8\std{0.35} & 46.5 \\
DAEL (\emph{ours}) & \textbf{70.8}\std{0.14} & {26.5}\std{0.13} & \textbf{57.4}\std{0.28} & 12.2\std{0.70} & \textbf{65.0}\std{0.23} & \textbf{60.6}\std{0.25} & \textbf{48.7} \\
\bottomrule
\end{tabular}
\label{tab:results_domainnet}
}
\\
\subfloat[miniDomainNet.]{
\tabstyle{9pt}
\begin{tabular}{l | c c c c | c}
\toprule
Method & Clipart & Painting & Real & Sketch & Avg \\
\midrule
Oracle & 72.59\std{0.30} & 60.53\std{0.74} & 80.47\std{0.34} & 63.44\std{0.15} & 69.26 \\ \midrule
Source-only & 63.44\std{0.76} & 49.92\std{0.71} & 61.54\std{0.08} & 44.12\std{0.31} & 54.76 \\
MCD~\cite{cvpr18MCD} & 62.91\std{0.67} & 45.77\std{0.45} & 57.57\std{0.33} & 45.88\std{0.67} & 53.03 \\
DCTN~\cite{xu2018deep} &62.06\std{0.60} & 48.79\std{0.52} & 58.85\std{0.55} & 48.25\std{0.32} & 54.49 \\
DANN~\cite{ganin2016domain} & 65.55\std{0.34} & 46.27\std{0.71} & 58.68\std{0.64} & 47.88\std{0.54} & 54.60 \\
M$^3$SDA~\cite{iccv19DomainNet} & 64.18\std{0.27} & 49.05\std{0.16} & 57.70\std{0.24} & 49.21\std{0.34} & 55.03 \\
MME~\cite{iccv19MME} & 68.09\std{0.16} & 47.14\std{0.32} & 63.33\std{0.16} & 43.50\std{0.47} & 55.52 \\
DAEL (\emph{ours}) & \textbf{69.95}\std{0.52} & \textbf{55.13}\std{0.78} & \textbf{66.11}\std{0.14} & \textbf{55.72}\std{0.79} & \textbf{61.73} \\
\bottomrule
\end{tabular}
\label{tab:results_minidn}
}
\end{table*}

\begin{table*}[t]
\caption{Comparing DAEL with state of the art on DG datasets PACS (left) and Office-Home (right).}
\label{tab:DG_results}
\tabstyle{3pt}
\begin{tabular}{l | c c c c | c | c c c c | c}
\toprule
\multirow{2}{*}{Method} & \multicolumn{5}{c|}{PACS} & \multicolumn{5}{c}{Office-Home} \\
\cmidrule(lr){2-6}\cmidrule(lr){7-11}
& Art & Cat & Pho & Skt & Avg & Art & Clp & Prd & Rel & Avg \\
\midrule
Vanilla & 77.0 & 75.9 & \textbf{96.0} & 69.2 & 79.5 & 58.9 & 49.4 & \textbf{74.3} & \textbf{76.2} & 64.7 \\
MMD-AAE~\cite{li2018mmdaae} & 75.2 & 72.7 & \textbf{96.0} & 64.2 & 77.0 & 56.5 & 47.3 & 72.1 & 74.8 & 62.7 \\
CCSA~\cite{motiian2017unified} & 80.5 & 76.9 & 93.6 & 66.8 & 79.4 & \textbf{59.9} & {49.9} & 74.1 & 75.7 & {64.9} \\
JiGen~\cite{cvpr19jigen} & 79.4 & 75.3 & \textbf{96.0} & 71.6 & 80.5 & 53.0 & 47.5 & 71.5 & 72.8 & 61.2 \\
CrossGrad~\cite{shankar2018generalizing} & 79.8 & 76.8 & \textbf{96.0} & 70.2 & 80.7 & 58.4 & 49.4 & 73.9 & 75.8 & 64.4 \\
Epi-FCR~\cite{li2019episodic} & 82.1 & {77.0} & 93.9 & 73.0 & 81.5 & - & - & - & - & - \\
DMG~\cite{chattopadhyay2020learning} & 76.9 & \textbf{80.4} & 93.4 & 75.2 & 81.5 & - & - & - & - & - \\
DAEL (\emph{ours}) & \textbf{84.6} & 74.4 & 95.6 & \textbf{78.9} & \textbf{83.4} & 59.4 & \textbf{55.1} & 74.0 & 75.7 & \textbf{66.1} \\
\bottomrule
\end{tabular}
\end{table*}

\begin{figure*}[t]
    \centering
    \includegraphics[width=.7\textwidth]{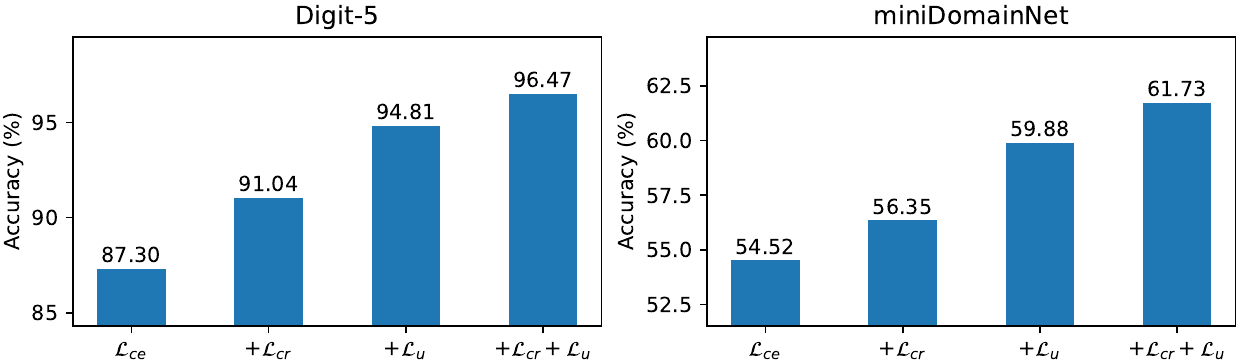}
    \caption{Ablation study for evaluating each component in Eq.~\eqref{eq:L_full}.}
    \label{fig:abl_sty}
\end{figure*}

The implementation details (of all experiments in this paper) are provided in Appendix~\ref{appx:imp_details}. To ensure the results are convincing, we run each experiment three times and report the mean accuracy and standard deviation. The results of baseline models are from either their papers (if reported) or our re-implementation (only when their source code is available).

\textbf{Results.}
Following the standard test protocol~\cite{iccv19DomainNet}, one domain is used as target and the rest as sources, and classification accuracy on the target domain test set is reported. Table~\ref{tab:msUDA_results} shows the results on the multi-source UDA datasets. We summarize our findings as follows.
(1) In terms of the overall performance (the rightmost Avg column), DAEL achieves the best results on all three datasets, outperforming the second-best methods by large margins: 3.51\% on Digit-5, 2.2\% on DomainNet and 6.21\% on miniDomainNet.

(2) On the small Digit-5 dataset, DAEL achieves near-oracle performance (our 96.47\% vs.~oracle's 97.00\%). In particular, MNIST-M and SVHN are the two most difficult domains as can be seen in Fig.~\ref{fig:examples_digit5_domainnet}---MNIST-M has complex backgrounds while SVHN contains blurred and cluttered digits. Those distinctive features make MNIST-M and SVHN drastically different from other domains and thus make the adaptation task harder. Nonetheless, DAEL obtains the highest accuracy which beats M$^3$SDA---the 2nd best method---by 11.62\% on MNIST-M and 4.06\% on SVHN.

(3) On the large-scale DomainNet/miniDomainNet, DAEL achieves the best performance among all methods. Notably, compared with the latest state of the art on DomainNet, i.e. CMSS, DAEL obtains a clear margin of 2.2\% on average, which demonstrates the advantage of exploiting complementarity between source domains for ensemble prediction.

(4) Compared with M$^3$SDA, the most related method to ours that also has domain-specific classifiers, DAEL is superior on all three datasets. This is because aligning distributions between the target and each individual source, as in M$^3$SDA, is difficult due to large domain variations between sources.

\subsection{Experiments on Domain Generalization}

\textbf{Datasets.}
(1) \textbf{PACS}~\cite{li2017deeper} is a commonly used DG dataset with four domains: Photo (1,670 images), Art Painting (2,048 images), Cartoon (2,344 images) and Sketch (3,929 images). There are seven object categories: dog, elephant, giraffe, guitar, horse, house and person.
(2) \textbf{Office-Home}~\cite{office_home} contains around 15,500 images of 65 categories, which are related to office and home objects. Similar to PACS, there are four domains: Artistic, Clipart, Product and Real World.
For evaluation, we follow the prior works~\cite{li2017deeper,cvpr19jigen,li2019episodic} to use the leave-one-domain-out protocol, i.e. choosing one domain as the (unseen) test domain and using the remaining three as source domains for model training.

\textbf{Results.}
The comparison with the state-of-the-art DG methods is shown in Table~\ref{tab:DG_results}. Overall, DAEL achieves the best results on both datasets with clear margins against all competitors. We provide a more detailed discussion as follows.
(1) DAEL is clearly better than the distribution alignment methods, i.e. CCSA and MMD-AAE, with $\geq$4\% improvement on PACS and $\geq$1.2\% improvement on Office-Home. This is not surprising because the distribution alignment theory~\cite{ben2010theory} developed for DA does not necessarily work for DG (which does not have access to target data).
(2) Compared with the recent self-supervised method JiGen, DAEL obtains a clear improvement of 2.9\% on PACS. The gap is further increased to 4.9\% on Office-Home. When it comes to CrossGrad, a state-of-the-art data augmentation method, DAEL achieves clear improvements as well.
(3) The recently proposed Epi-FCR shares a similar design choice with DAEL---to simulate domain shift during training. Again, DAEL is clearly superior thanks to the design of collaborative ensemble learning. Further, Epi-FCR requires domain-specific feature extractors, as well as additional domain-agnostic feature extractors and classifiers, incurring much higher computational cost.

\subsection{Analysis}

\begin{table*}[t]
    \caption{Evaluation of design choices in DAEL. D-5: Digit-5. miniDN: miniDomainNet.}
    \label{tab:design_choices}
    \centering
    \subfloat[Collaborative ensemble vs.~individual expert learning.]{
    \tabstyle{3pt}
    \begin{tabular}{l|cc}
    \toprule
    & D-5 & miniDN \\
    \midrule
    Col. & 96.47 & 61.73 \\
    Ind. & 93.07 & 60.20 \\
    \bottomrule
    \end{tabular}
    \label{tab:col_vs_ind}
    }
    ~
    \subfloat[Learning an ensemble of classifiers vs.~a single classifier.]{
    \tabstyle{3pt}
    \begin{tabular}{l|cc}
    \toprule
    & D-5 & miniDN \\
    \midrule
    Ensemble & 96.47 & 61.73 \\
    Single & 95.02 & 60.09 \\
    \bottomrule
    \end{tabular}
    \label{tab:ensemble_vs_single}
    }
    ~
    \subfloat[Expert's prediction vs.~real label (see $\mathcal{L}_{cr}$).]{
    \tabstyle{3pt}
    \begin{tabular}{l|cc}
    \toprule
    & D-5 & miniDN \\
    \midrule
    $E_i$ & 91.04 & 56.35 \\
    $Y$ & 90.84 & 56.11 \\
    \bottomrule
    \end{tabular}
    \label{tab:Lcr_target_expert_vs_real}
    }
    ~
    \subfloat[Expert's prediction vs.~ensemble prediction (see $\mathcal{L}_{u}$).]{
    \tabstyle{3pt}
    \begin{tabular}{l|cc}
    \toprule
    & D-5 & miniDN \\
    \midrule
    $E_{i}$ & 96.47 & 61.73 \\
    $\bar{E}$ & 95.86 & 60.16 \\
    \bottomrule
    \end{tabular}
    \label{tab:Lu_target_expert_vs_ensemble}
    }
\end{table*}

\textbf{Ablation study.}
We start from the baseline ensemble model trained by $\mathcal{L}_{ce}$ only and progressively add $\mathcal{L}_{cr}$ and $\mathcal{L}_{u}$ (see Eq.~\eqref{eq:L_full}). The results are shown in Fig.~\ref{fig:abl_sty}. Each of $\mathcal{L}_{cr}$ and $\mathcal{L}_{u}$ contributes positively to the performance. Combining $\mathcal{L}_{cr}$ and $\mathcal{L}_{u}$ gives the best performance, which confirms their complementarity.

\begin{figure}[t]
    \centering
    \includegraphics[width=\columnwidth]{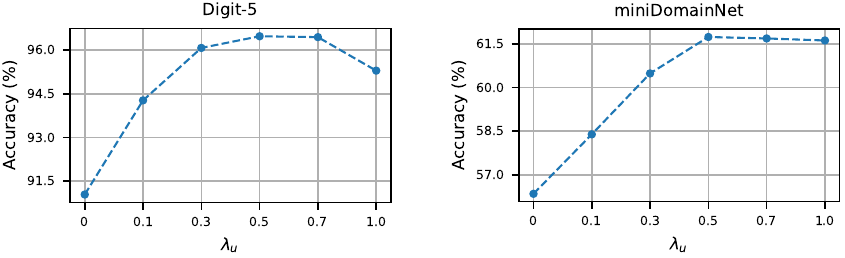}
    \caption{Sensitivity of $\lambda_{u}$.}
    \label{fig:effect_hyperparams}
\end{figure}

\textbf{Sensitivity of $\lambda_u$.}
Fig.~\ref{fig:effect_hyperparams} shows that the performance soars from $\lambda_u=0$ to $\lambda_u=0.5$ and remains relatively stable between $\lambda_u=0.5$ and $\lambda_u=1.0$. The overall results suggest that the model's performance is in general insensitive to $\lambda_{u}$ around $0.5$.

\textbf{Collaborative ensemble or individual expert training?}
As discussed in the gradient analysis part in Methodology, collaborative learning aggregates gradients from different experts, which can better exploit the complementarity between different sources. We justify this design in Table~\ref{tab:col_vs_ind} where collaborative learning shows clear improvements over individual learning.

\textbf{Learning an ensemble of classifiers or a single classifier?}
The motivation for the former is to enable the model to better handle complicated source data distributions---as discussed before, learning a single classifier forces the model to erase domain-specific knowledge that could otherwise be useful for recognition in the target domain. To justify this design, we switch from the ensemble classifiers to a single classifier while keeping other designs unchanged. Table~\ref{tab:ensemble_vs_single} confirms that learning an ensemble of classifiers is essential.

\textbf{Using expert's prediction or real label in $\mathcal{L}_{cr}$?}
Table~\ref{tab:Lcr_target_expert_vs_real} shows that using real label ($Y$) is slightly worse than using expert's prediction ($E_i$). This is because expert's prediction automatically encodes the relations between classes (reflected in the soft probability distribution~\cite{wu2018unsupervised}), thus providing better supervisory signal.

\textbf{Most confident expert's prediction vs.~ensemble prediction for $\mathcal{L}_u$?}
Table~\ref{tab:Lu_target_expert_vs_ensemble} suggests that using the most confident expert's output ($E_{i^*}$) is better. A plausible explanation is that ensembling smooths out the overall probability distribution when experts have disagreements, which may lead to potentially correct instances discarded due to weak confidence (i.e.~probability less than the confidence threshold $\epsilon$).

\begin{figure*}[t]
    \centering
    \includegraphics[width=\textwidth]{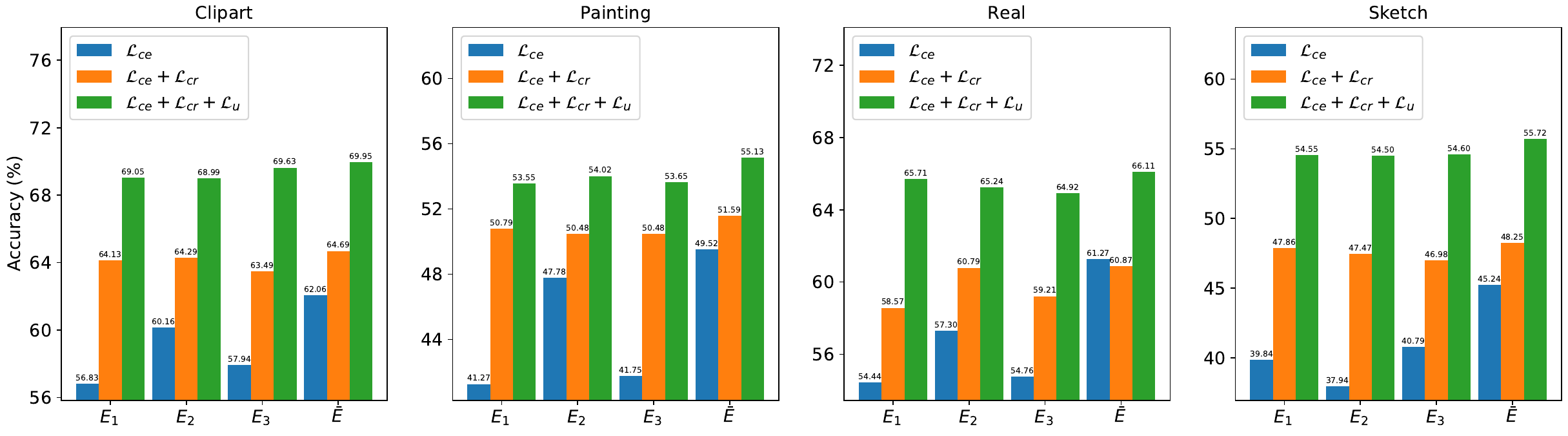}
    \caption{Individual experts ($E_{1-3}$) vs.~ensemble ($\bar{E}$) on miniDomainNet.}
    \label{fig:expert_vs_ensemble}
\end{figure*}

\begin{figure*}[t]
    \centering
    \includegraphics[width=\textwidth]{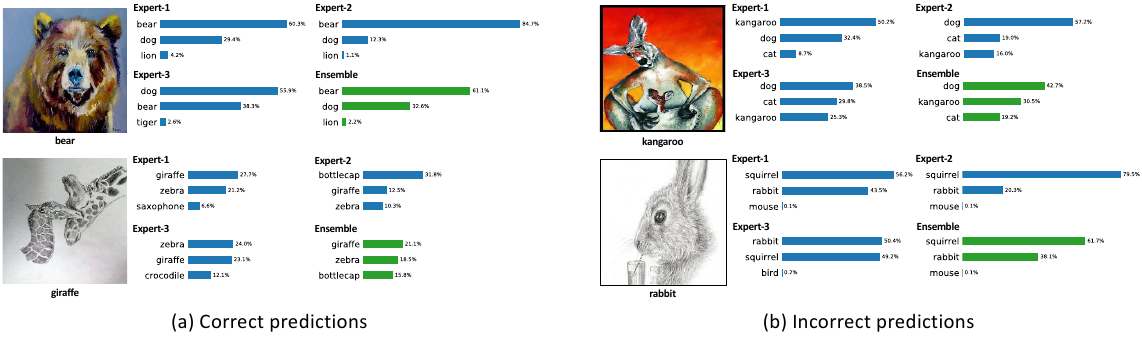}
    \caption{Visualization of predicted classes (top-3) and the corresponding confidence by each expert and the ensemble.}
    \label{fig:vis_pred}
\end{figure*}

\textbf{Diagnosis into individual experts.}
Fig.~\ref{fig:expert_vs_ensemble} shows the performance of each individual source expert versus the ensemble on miniDomainNet trained with different losses. For $\mathcal{L}_{ce}$ (blue bars), the variance between $E_{1-3}$ is large and each individual's performance is low, indicating that the experts are themselves biased (overfitting). Comparing $\mathcal{L}_{ce}+\mathcal{L}_{cr}$ (orange bars) with $\mathcal{L}_{ce}$, we observe that the variance between $E_{1-3}$ is reduced and each individual's performance is significantly improved, leading to a much stronger ensemble. By adding $\mathcal{L}_{u}$ (green bars), each individual's performance is further boosted, and hence the ensemble. To better understand how the ensemble helps prediction, we visualize the top-3 classes predicted by each expert and the ensemble in Fig.~\ref{fig:vis_pred}. In Fig.~\ref{fig:vis_pred}(a) top, expert-3 mis-recognizes the bear as dog but the ensemble prediction is dominated by the correct predictions made by expert-1 and -2. A similar pattern can be observed in Fig.~\ref{fig:vis_pred}(a) bottom. Fig.~\ref{fig:vis_pred}(b) provides examples of incorrect predictions where we observe that the model struggles to differentiate between classes of similar features. For example, the kangaroo in Fig.~\ref{fig:vis_pred}(b) top looks indeed similar to a dog due to the similarity in the face and skin texture---without looking at the body structure and the pouch. Such mistakes might be avoided by using neural networks that can extract features at multiple scales~\cite{zhou2019omni}.

\begin{table}[t]
    \caption{Comparison with baseline models trained using strong augmentation on miniDomainNet.}
    \label{tab:baseline_with_straug}
    \centering
    \tabstyle{3pt}
    \begin{tabular}{l | ccccc}
    \toprule
    Model & Clipart & Painting & Real & Sketch & Avg \\
    \midrule
    Source-only & 63.44 & 49.92 & 61.54 & 44.12 & 54.76 \\
    Source-only+$A(\cdot)$ & 64.34 & 49.31 & 58.81 & 48.02 & 55.12 \\
    MSDA & 64.18 & 49.05 & 57.70 & 49.21 & 55.03 \\
    MSDA+$A(\cdot)$ & 65.08 & 49.68 & 57.04 & 54.29 & 56.52 \\
    MME & 68.09 & 47.14 & 63.33 & 43.50 & 55.52 \\
    MME+$A(\cdot)$ & 68.25 & 49.21 & 60.26 & 44.23 & 55.49 \\
    DAEL & \textbf{69.95} & \textbf{55.13} & \textbf{66.11} & \textbf{55.72} & \textbf{61.73} \\
    \bottomrule
    \end{tabular}
\end{table}

\textbf{Augmentation strategy.}
Recall that we use weak and strong augmentations for pseudo-label generation and prediction respectively. The rationales behind this design are: 1) We need the expert to provide accurate supervision (pseudo labels) to the non-expert ensemble so we feed the expert with weakly augmented data; 2) We apply strong augmentation to the non-expert ensemble in order to reduce overfitting to noisy pseudo labels. If we swap these two augmentations, i.e.~using weak augmentation for prediction while strong augmentation for pseudo-label generation, the accuracy decreases from 61.73\% to 54.32\% on miniDomainNet. In addition, if strong augmentation is applied to both branches, the accuracy declines to 58.05\%, which suggests the weak-strong augmentation strategy is essential. These observations have also been reported by Sohn et al.~\cite{sohn2020fixmatch}.

To justify that using strong augmentation is not the sole contributor to our approach, we compare with top-performing baselines that also use strong augmentation in Table~\ref{tab:baseline_with_straug}. We can see that the improvements brought by strong augmentation for the baselines are rather limited, and the gap with our method remains large. This result confirms that collaborative ensemble learning is the key to our superior performance.

\begin{figure}[t]
    \centering
    \includegraphics[width=\columnwidth]{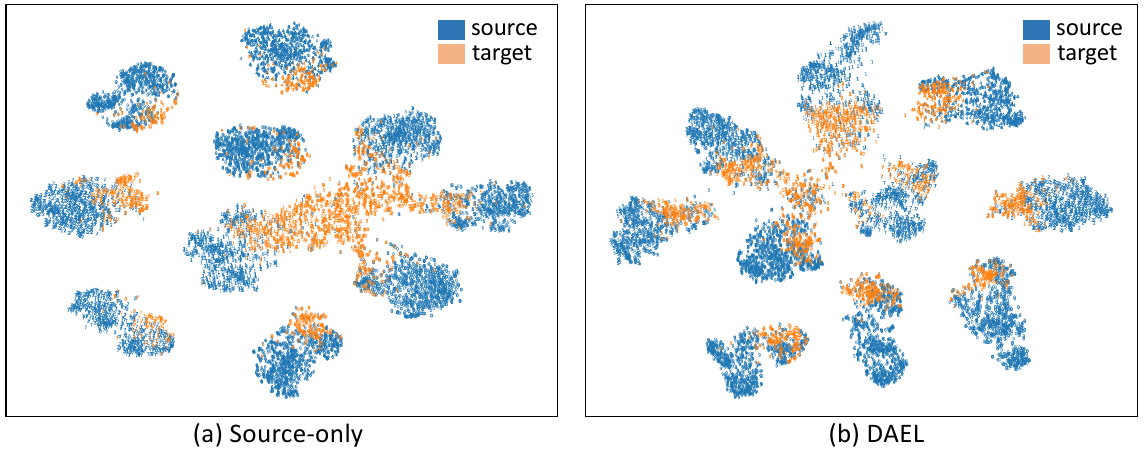}
    \caption{Visualization of features from Digit-5 using t-SNE~\cite{tsne}.}
    \label{fig:tsne}
\end{figure}

\textbf{Visualization of features.}
We use t-SNE~\cite{tsne} to visualize the features learned by Source-only and our DAEL. Figure~\ref{fig:tsne} shows that the target features learned by Source-only are poorly aligned---the model cannot clearly differentiate between ``1'', ``0'', ``5'', ``3'', and ``8'' (zoom-in to see the class labels). In contrast, the target features learned by DAEL have a much smaller domain discrepancy with the source features and exhibit clearer class-based clustering patterns.

\section{Conclusion}
Our approach, domain adaptive ensemble learning (DAEL), takes the first step toward a general framework for generalizing neural networks from multiple source domains to a target domain. When target data are not provided (the DG problem), DAEL shows promising out-of-distribution generalization performance on PACS and Office-Home. When unlabeled target data are accessible (the UDA problem), DAEL leverages pseudo labels and follows the same collaborative ensemble learning strategy as used in the DG setting to prompt the emergence of domain-generalizable features. Currently, to avoid overfitting to noisy pseudo labels, advanced data augmentation methods are used. However, the design of data augmentation is often task-specific, e.g., for digit recognition we cannot use random flip; for fine-grained recognition, color distortion might be discarded. Future work can focus on new algorithmic designs to mitigate the overfitting problem in a more flexible way.


%

\appendices
\section{Implementation Details} \label{appx:imp_details}
\textbf{Experiments on domain adaptation.}
SGD with momentum is used as the optimizer, and the cosine annealing rule~\cite{cosineLR} is adopted for learning rate decay. For Digit-5, the CNN backbone is constructed with three convolution layers and two fully connected layers~\cite{iccv19DomainNet}. For each mini-batch, we sample from each domain 64 images. The model is trained with an initial learning rate of 0.05 for 30 epochs. For DomainNet, we use ResNet101~\cite{he2016deep} as the CNN backbone and sample from each domain 6 images to form a mini-batch. The model is trained with an initial learning rate of 0.002 for 40 epochs. For miniDomainNet, we use ResNet18~\cite{he2016deep} as the CNN backbone. Similarly, we sample 64 images from each domain to form a mini-batch and train the model for 60 epochs with an initial learning rate of 0.005. For all UDA experiments, we set $\lambda_u=0.5$ in all datasets. The sensitivity of $\lambda_u=0.5$ to performance is investigated in Fig.~\ref{fig:effect_hyperparams}.

\textbf{Experiments on domain generalization.}
ResNet18 is used as the CNN backbone as in previous works~\cite{cvpr19jigen,li2019episodic}. SGD with momentum is used to train the model for 40 epochs with an initial learning rate of 0.002. The learning rate is further decayed by the cosine annealing rule. Each mini-batch contains 30 images (10 per source domain). Note that the $\mathcal{L}_u$ term in Eq.(4) is discarded here as no target data is available for training.

\section{Pseudo-Code} \label{appx:pseudo_code}
The full algorithm of domain adaptive ensemble learning is presented in Alg.~\ref{alg:one_step_loss}.

\begin{algorithm}[h]
\caption{Pseudo-code for loss computation in DAEL.}
\label{alg:one_step_loss}
\footnotesize
\begin{algorithmic}[1] 
  \STATE \textbf{Require}: labeled source mini-batches $\{ (X^i, Y^i) \}_{i=1}^K$, unlabeled target mini-batch $X^t$, source experts $\{ E_i \}_{i=1}^K$, weak/strong augmentation $a(\cdot)$/$A(\cdot)$, hyper-parameter $\lambda_{u}$.
  \STATE \textbf{Return}: loss $\mathcal{L}$.
  \STATE $\mathcal{L}_{ce} = 0$ \hspace{1em} \codecmt{Initialize $\mathcal{L}_{ce}$}
  \STATE $\mathcal{L}_{cr} = 0$ \hspace{1em} \codecmt{Initialize $\mathcal{L}_{cr}$}
  \FOR{$i = 1$ to $K$}
    \STATE \codecmt{Domain-specific expert learning}
    \STATE $\tilde{X}^i = a(X^i)$ \hspace{1em} \codecmt{Apply weak augmentation to $X^i$}
    \STATE $\tilde{Y}^i = E_i (\tilde{X}^i)$ \hspace{1em} \codecmt{Compute prediction for expert-i}
    \STATE $\mathcal{L}_{ce} = \mathcal{L}_{ce} + \mathrm{CrossEntropy}(\tilde{Y}^i, Y^i)$ \hspace{1em} \codecmt{Compute cross-entropy loss for expert-i}
    \STATE \codecmt{Collaborative ensemble learning for source data}
    \STATE $\hat{X}^i = A(X^i)$ \hspace{1em} \codecmt{Apply strong augmentation to $X^i$}
    \STATE $\hat{Y}^i = \frac{1}{K-1} \sum_{j \neq i} E_j (\hat{X}^i)$ \hspace{1em} \codecmt{Compute ensemble prediction of non-experts}
    \STATE $\mathcal{L}_{cr} = \mathcal{L}_{cr} + \mathrm{MSE}(\hat{Y}^i, \tilde{Y}^i)$ \hspace{1em} \codecmt{Compute consistency loss for non-experts}
  \ENDFOR
  \STATE $\mathcal{L}_{ce} = \mathcal{L}_{ce} / K$
  \STATE $\mathcal{L}_{cr} = \mathcal{L}_{cr} / K$
  \STATE $\mathcal{L} = \mathcal{L}_{ce} + \mathcal{L}_{cr}$
  \IF{$X^t$ is available}
    \STATE \codecmt{Collaborative ensemble learning for unlabeled target data}
    \STATE $\tilde{X}^t = a(X^t)$ \hspace{1em} \codecmt{Apply weak augmentation to $X^t$}
    \STATE $\tilde{Y}^t, M = \mathrm{PseudoLabel} (\{ E_i (\tilde{X}^t) \}_{i})$ \hspace{1em} \codecmt{Get pseudo labels and instance masks}
    \STATE $\hat{X}^t = A(X^t)$ \hspace{1em} \codecmt{Apply strong augmentation to $X^t$}
    \STATE $\hat{Y}^t = \frac{1}{K} \sum_{i} E_i (\hat{X}^t)$ \hspace{1em} \codecmt{Compute ensemble prediction of all experts}
    \STATE $\mathcal{L}_{u} = \mathrm{CrossEntropy}(\hat{Y}^t, \tilde{Y}^t, M)$ \hspace{1em} \codecmt{Compute cross-entropy loss for all experts}
    \STATE $\mathcal{L} = \mathcal{L} + \lambda_{u} \mathcal{L}_{u}$
  \ENDIF
\end{algorithmic}
\end{algorithm}




\ifCLASSOPTIONcaptionsoff
  \newpage
\fi



\bibliographystyle{IEEEtran}
\bibliography{reference}
\end{document}